\pdfoutput=1

\documentclass[11pt]{article}

\usepackage{acl}

\usepackage{times}
\usepackage{latexsym}
\usepackage{tikz-dependency}

\usepackage{graphicx}
\usepackage{multirow}
\usepackage{svg}
\usepackage{amsmath}

\usepackage{color}
\usepackage{transparent}

\usepackage[T1]{fontenc}

\usepackage[utf8]{inputenc}

\usepackage{microtype}

\title{Multipath parsing in the brain}

\author{Berta Franzluebbers \\
  University of Georgia \\
  \texttt{berta@uga.edu} \\\And
  Donald Dunagan \\
  University of Georgia \\
  \texttt{dgd45125@uga.edu} \\\And
  Milo\v{s} Stanojevi\'{c} \\
  Google DeepMind \\
  \texttt{stanojevic@google.com} \\\AND
  Jan Buys \\
  University of Cape Town \\
  \texttt{jbuys@cs.uct.ac.za} \\\And
  John T. Hale \\
  University of Georgia \\
  \texttt{jthale@uga.edu} \\}

\begin{document}

\maketitle

\begin{abstract}
Humans understand sentences word-by-word, in the order that they hear them. %
This incrementality entails resolving temporary ambiguities about syntactic~relationships.
We investigate how humans process these syntactic ambiguities by correlating
predictions from incremental generative dependency parsers with timecourse~data from
people undergoing functional neuroimaging while listening to an audiobook.
In particular, we compare competing hypotheses regarding the number of developing syntactic~analyses in play during word-by-word comprehension: one vs more than one.
This comparison involves evaluating syntactic~surprisal from a state-of-the-art dependency parser with LLM-adapted encodings
against an existing fMRI dataset. In both English and Chinese data, we find evidence for multipath parsing.
Brain regions associated with this multipath effect include bilateral superior temporal gyrus.
\end{abstract}

\section{Introduction}

\begin{figure}[t!]
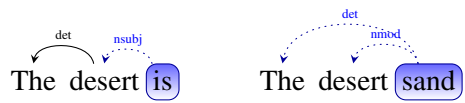

\setlength{\belowcaptionskip}{-18pt}
    \centering
    \begin{dependency}[theme=simple]
        \tikzstyle{word}=[draw=blue!60!black, shade, text=black, top color=blue!60, rounded corners]
        \begin{deptext}
        The \& desert \& |[word]| is  \\
        \end{deptext}
        \depedge{2}{1}{det}
        \depedge[edge style={blue!60!black,dotted},label style={text=blue}]{3}{2}{nsubj}
    \end{dependency}
    \hspace{3ex}
    \begin{dependency}[theme=simple]
        \tikzstyle{word}=[draw=blue!60!black, shade, text=black, top color=blue!60, rounded corners]
        \begin{deptext}
        The \& desert \& |[word]| sand  \\
        \end{deptext}
        \depedge[edge style={blue!60!black,dotted},label style={text=blue}]{3}{1}{det}
       \depedge[edge style={blue!60!black,dotted},label style={text=blue}]{3}{2}{nmod}
   \end{dependency}
    \caption{A sentence fragment from our stimulus~text showing temporary syntactic ambiguity about the correct relationship between \textit{The} and \textit{desert}, which is resolved by hearing the next word (blue).}
    \label{fig:ex_ambiguous_dep_graph}
\end{figure}

A major unsolved problem in computational~psycholinguistics is determining whether human sentence comprehension considers a \emph{single} analysis~path\footnote{The ``paths'' terminology (vs ``serial'' or ``parallel'')
serves to categorize the cognitive~issue as one of process rather than architecture \citep[see][\S3.3.3]{lewis:arch}.} at a time, or whether it sometimes entertains \emph{multiple} lines of reasoning about the structure of a sentence. %
These lines of reasoning typically correspond to syntactic ambiguities (e.g. between Subject and Modifier as shown in Figure~\ref{fig:ex_ambiguous_dep_graph}).
Multipath accounts of ambiguity resolution \citep[e.g.][]{gibson91,jurafsky1996probabilistic}
are quite different from single-path accounts \citep[e.g.][]{frazier78,marcus80}.  %
The debate within cognitive~science has focused on the special case of garden~path sentences
leaving~open the broader question: Is human~comprehension ever multipath in the sense of Ranked~Parallel parsing? If it is, where in the brain does such multipath~parsing happen?

We take~up the call to answer this question \citep[e.g.][]{lewis00,gibsonpearlmutter00}
by combining broad~coverage incremental~parsing \citep[e.g.][]{nivre2004incrementality} with information-theoretical complexity metrics to derive predictions about neuroimaging.  Analyzing an entire book \citep{li2022petit}  makes it possible to examine -- for the first time --  all of the ambiguities that are implicated by a given conception of syntax and a given parsing strategy -- not just those attested in classic garden path materials \citep{mason2003ambiguity,hopf03,rodd2010functional}.

We extend the state of the art in incremental generative dependency parsing by using a large language model \cite[BLOOM;][]{le-scao-etal-2022-language} together with parameter efficient fine-tuning using adapters~\cite{pfeiffer2020AdapterHub}.
We fix the parsing~strategy while varying the number of allowable analysis~paths %
and connect %
neural~data with parser~actions via surprisal \citetext{for a review of this information-theoretical metric see \citealp{hale2016-dm} or neuroimaging~studies such as \citealp{brennan2016-tv,Henderson:2016yq,shain2020fmri,brennan:2020ku}}.
This sets~up a contrast between a multipath model that considers (at most) five~paths at a time, versus a single-path model that can only consider one at a time.
The results, reported in section~\ref{sec:results}, ultimately support the multipath~view, as the surprisals from the five-way parser are better-correlated with the neuroimaging data. This obtains for both English and Chinese.

\begin{table*}
    \centering
    \begin{tabular}{|c|c|c|c|c|}
        \hline
        Action & Before & After & Arc & Probability  \\
        \hline
        Shift & ($\sigma | i , j$) & ($\sigma|i|j , j+1$) & - & $p_{tr}(\mathrm{sh}|h_i,h_j)p_{gen}(w_{j+1}|h_i,h_j)$\\
        Left-arc & ($\sigma | i , j$) & ($\sigma , j$) & $j \rightarrow i$ & $p_{tr}(\mathrm{re}|h_i,h_j)p_{dir}(\mathrm{la}|h_i,h_j)$\\
        Right-arc & ($\sigma |l| i , j$) & ($\sigma|l , j$) & $l \rightarrow i$ & $p_{tr}(\mathrm{re}|h_i,h_j)p_{dir}(\mathrm{ra}|h_i,h_j)$\\
        \hline
    \end{tabular}
    \caption{The arc-hybrid transition system of \citet{kuhlmann2011dynamic} defines the possible parser actions (shift, left-arc, and right-arc) as transitions from previous to current parser states. States are indicated by (stack, current index) tuples, e.g. $w_i$ is the token on top of the stack ($\sigma$), and $w_j$ is the current index token at time step $j$. The probabilities associated with these actions are decomposed into complementary shift (sh) and reduce (re) transitions, and complementary left-arc (la) and right-arc (ra) arc directions. The shift action also includes the probability of generating the next token $w_{j+1}$.}
    \label{tab:transitions}
\end{table*}

\section{Related Work}

This paper follows a line of research which aims to characterize human language processing by evaluating word-by-word difficulty predictions against neuroimaging data from the brain.
\citet{hale2022neurocomputational} reviews many studies that fit into this tradition.
Most of them consider just a single gold~standard or single system-assigned analysis. 
\citet{hale-etal-2018-finding} and \citet{crabbe-etal-2019-variable} are notable exceptions to this general trend
because they derive predictions from multiple analyses that would be considered as part of a beam. %
Hale and Crabb\'{e} work with phrase~structure. By contrast, the present~study uses dependency parsing.
This choice is motivated by prior~work in neurolinguistics \cite[e.g.][]{adm:30}.
For instance, \citet{jixing:minparse} relate a dependency-based structural distance~metric
to hemodynamic~activity in
left posterior temporal~lobe, among other brain areas.
Their study used spoken English materials.
\citet{lopopolo21} apply a related metric to Dutch.  \citet{oota-etal-2023-brain} use graph neural networks
to embed dependency~analyses of sentence-initial substrings.
Analyzing brain~responses to written, rather than spoken stimuli,
via the encoding approach of \citet{reddy21},
\citeauthor{oota-etal-2023-brain} identify many of the same brain~areas as the studies mentioned above.

The single-path vs multipath question is itself motivated by prior~work with eyetracking data.
By varying the number of paths available to a parser,
\citet{boston2011parallel} find support for Ranked Parallel parsing \citetext{in the sense of
\citealt{gibson91} and \citealt{jurafsky1996probabilistic}, further characterized below in \S\ref{sec:rankedparallel}}.

Apart from these cognitive considerations, a complementary motivation for this research is to investigate the parsing improvements
that come by leveraging large language models (LLMs; see accuracy scores in Table~\ref{tab:parser_accuracy}).
Such models are trained on datasets that far exceed classic treebanks in size. This allows LLMs to capture distributional regularities at a vast scale.
However, it is challenging to explain at an algorithmic level what these models are doing.

Instead of deriving a processing complexity metric directly from the output of a large language model,
or correlating LLM internal representations with fMRI data, as in \citet{schrimpf2020artificial} and \citet{caucheteux22},
this work uses initial substring encodings from an LLM to inform an incremental parser.
Our project also differs from \citet{eisape-etal-2022-probing},
who probe LLM~representations with a view towards inferring a single, unlabelled dependency~analysis.
In contrast, we use LLM~representations to score the set of all possible alternative analyses.
Our scientific~goal is to relate such parser~states to human brain~states, as observed via neuroimaging.

\section{Dependency Parsing}

\subsection{Parser Architecture}

The construction of our dependency parser relies on the work of \citet{buys2018neural}, which employs the transition-based parsing system detailed in \citet{nivre2008algorithms} and exactly enumerates all parser paths. Following \citet{kuhlmann2011dynamic}, dynamic programming is used to sum over all paths. We use the arc-hybrid transition system shown in Table~\ref{tab:transitions}
because it showed good parsing performance in \citet{buys2018neural}.
The generative parser assigns a probability to each of the words in the sentence incrementally, like any other language~model,
in contrast to discriminative~parsers which only predict transition~actions. 
The \textit{shift}~action predicts the next~word on the buffer, which corresponds to the \textit{buffer-next} model in \citet{buys2018neural}.

We update \citeauthor{buys2018neural} by encoding sequences not with an LSTM but with representations from the BigScience Large Open-science Open-access Multilingual Language Model  \citep[BLOOM;][]{le-scao-etal-2022-language}, chosen based on its open-access status and training on both English and Chinese data.  %
These BLOOM representations encode an entire initial substring up to and including a particular word (or subword token). 
A set~of~classifiers over these substring representations %
estimates the probability of each transition, word and arc~label. 
The classifiers correspond to the subscripted probabilities %
in the rightmost column of Table~\ref{tab:transitions}.

We use the pre-trained BLOOM-560 model. Larger BLOOM models showed no significant increase in accuracy. 
This corroborates
\citet{oh2023does} and \citet{pasquiou2023information}, who  %
find that smaller~models provide an equal (or better) fit to human~data.

Instead of fine-tuning the entire pre-trained BLOOM model on dependency parsing, we apply a Pfeiffer adapter after each layer \citep{pfeiffer2020AdapterHub}. This bottleneck~adapter introduces a new linear~layer which reduces the dimension from 1024 down to 64 and back up to 1024, for input into the next pre-trained BLOOM layer.
The adapter approach enables parameter-efficient fine-tuning while preserving the original language modelling representations as much as possible in the generative parser. 

The goal of these design~choices is to deliver linguistically plausible dependency analyses of sentence-initial substrings. Some aspects of the overall architecture play the role of auxiliary hypotheses that do not map on to the brain.
As subsection~\ref{sec:complexitymetric} lays~out in further detail,
the cognitive~claim is limited to proposals that (a) human~parsing involves recognizing dependency~relations via a schema like Table~\ref{tab:transitions}
and (b) surprising parser actions, within this schema, call for greater hemodynamic resources than do unsurprising ones.
Proposal~(a) is situated at Marr's middle level of analysis~\citep{marr:vision}.

\begin{figure}[h]
\setlength{\belowcaptionskip}{-12pt}
    \centering
    \includeinkscape[width=\columnwidth]{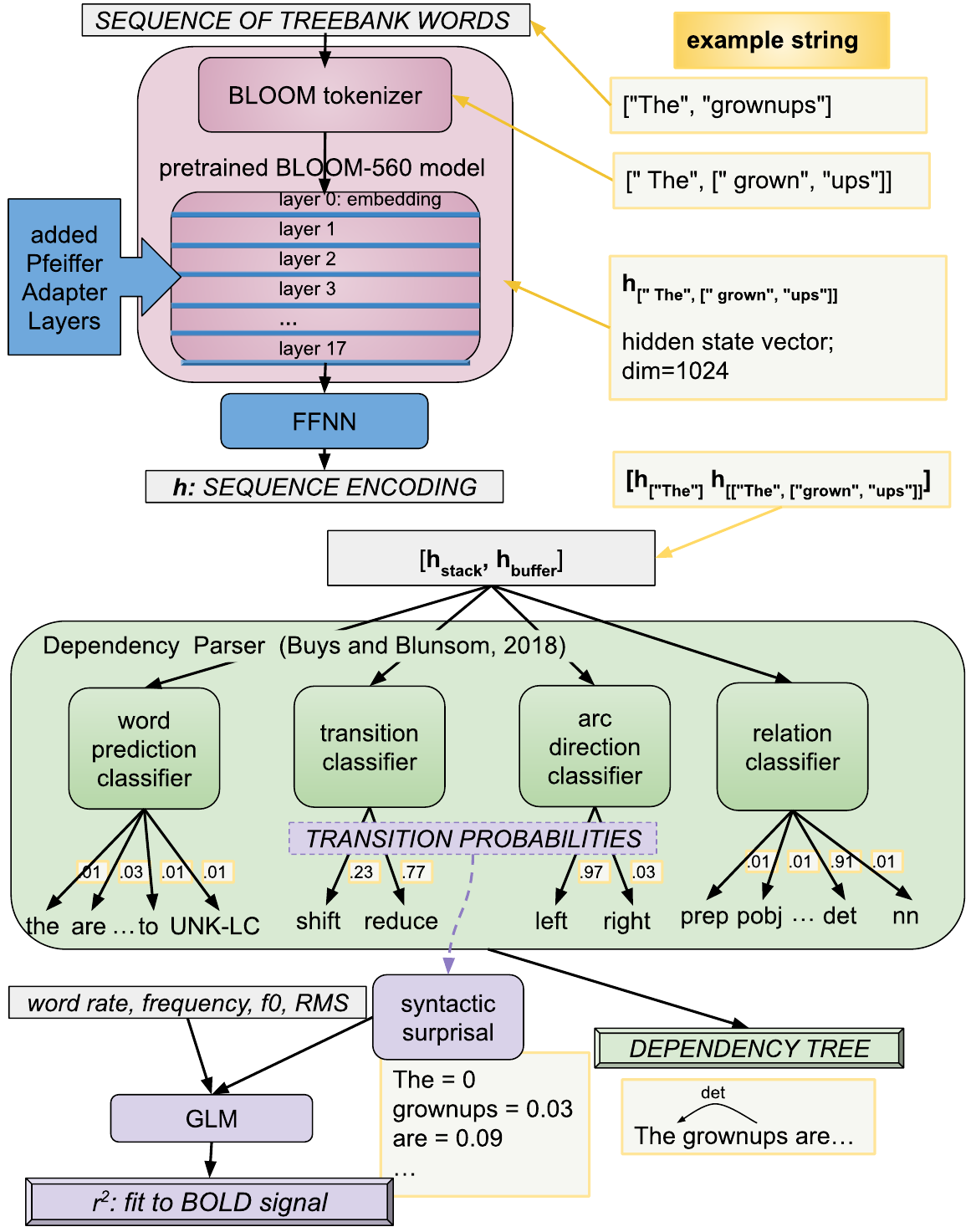}
    \caption{Pipeline Overview: Pretokenized text from the treebank corpora is pre-processed by adding a leading space to each token. The pre-trained BLOOM model is fine-tuned using the Pfeiffer Adapter, and an additional single-layer feedforward neural network completes the encoding step. The dependency parser takes tuples of sequence encodings (stack ($\sigma$), buffer ($\beta$)) as input, and its system of classifiers is trained from random weights. Note that the word prediction classifier predicts over the size of the training data vocabulary + Berkeley unknown tokens instead of the BLOOM vocabulary. The final dependency tree output is evaluated in Table \ref{tab:parser_accuracy}. After training, we use the parser's transition probabilities (see Table \ref{tab:transitions}) to calculate syntactic surprisal (Section \ref{sec:complexitymetric}), which is added to other regressors (see Section \ref{sec:Statistical}), in a General Linear Model (GLM) which predicts BOLD signal (Section \ref{sec:results}).}
    \label{fig:tokenization_flowchart}
\end{figure}

\subsection{Parser Training}
\label{sec:parser_training}

The parser is trained on English and Chinese treebanks annotated with Stanford Dependencies \cite{de2008stanford}. The English model is trained on the Penn~Treebank (PTB) version~3, with the standard~split of training on sections 02-21, development on section 22, and testing on section 23. The Chinese model is trained on  Chinese~Treebank version~7, with the standard split of files 0-2082 for training, development on files 2083-2242, and testing on files 2243-2447. This results in 39,832 training sentences for English and 19,457 training sentences for Chinese. 

\begin{table*}
    \centering
    \scalebox{1}{
    \begin{tabular}{|c|c|c c|c c|}
        \hline
        \multirow{2}{*}{Corpus} & \multirow{2}{*}{Model} & \multicolumn{2}{c|}{Dev} & \multicolumn{2}{c|}{Test}  \\
        & & LAS & UAS & LAS & UAS  \\
        \hline
        PTB-3 & \citet{buys2018neural} & 88.66 & 91.19 & 88.54 & 91.01 \\
        PTB-3 & English BLOOM (brain analysis) & 90.26 & 92.71 &  90.32 & 92.62 \\
        \hline
        CTB-7 & Chinese BLOOM (brain analysis) & 77.07 & 83.65 & 74.71 & 81.66 \\
        \hline
    \end{tabular}}
    \caption{Labeled attachment score (LAS) and unlabeled attachment score (UAS) for the English PTB corpus and Chinese CTB corpus}
    \label{tab:parser_accuracy}
\end{table*}

The parser's next word prediction classifier \mbox{predicts} over a limited vocabulary, where words seen only once in the training data are replaced with unknown word tokens according to the rules in the Berkeley Parser. 
This results in 23,830 word types for English and 19,671 for Chinese.
The LLM encoder, however, uses the BLOOM \mbox{sub-word} \mbox{tokenizer} and the BLOOM \mbox{vocabulary}, which is of size 250,680.\footnote{This vocabulary size mismatch
is one reason we did not directly assign next-word probabilities using the LLM.}

Token sequences are encoded by the LLM, selecting the encoding of the right-most sub-word of each word for parser predictions. 
This allows sequence encodings to be as detailed as possible while still limiting the vocabulary during training.
It should be noted, though, that using this encoding for words that are unknown to the parser's next word classifier prevents the parser from being strictly generative.  %
More information on strictly generative models can be found in the appendix.

While training, sentences are shuffled at each epoch, and within batches (batch size = 16), which are created from sentences of the same length. The BLOOM-560 model has 24 hidden layers, after encoding. 

\begin{figure}[h]
    \centering
    \includeinkscape[width=\columnwidth]{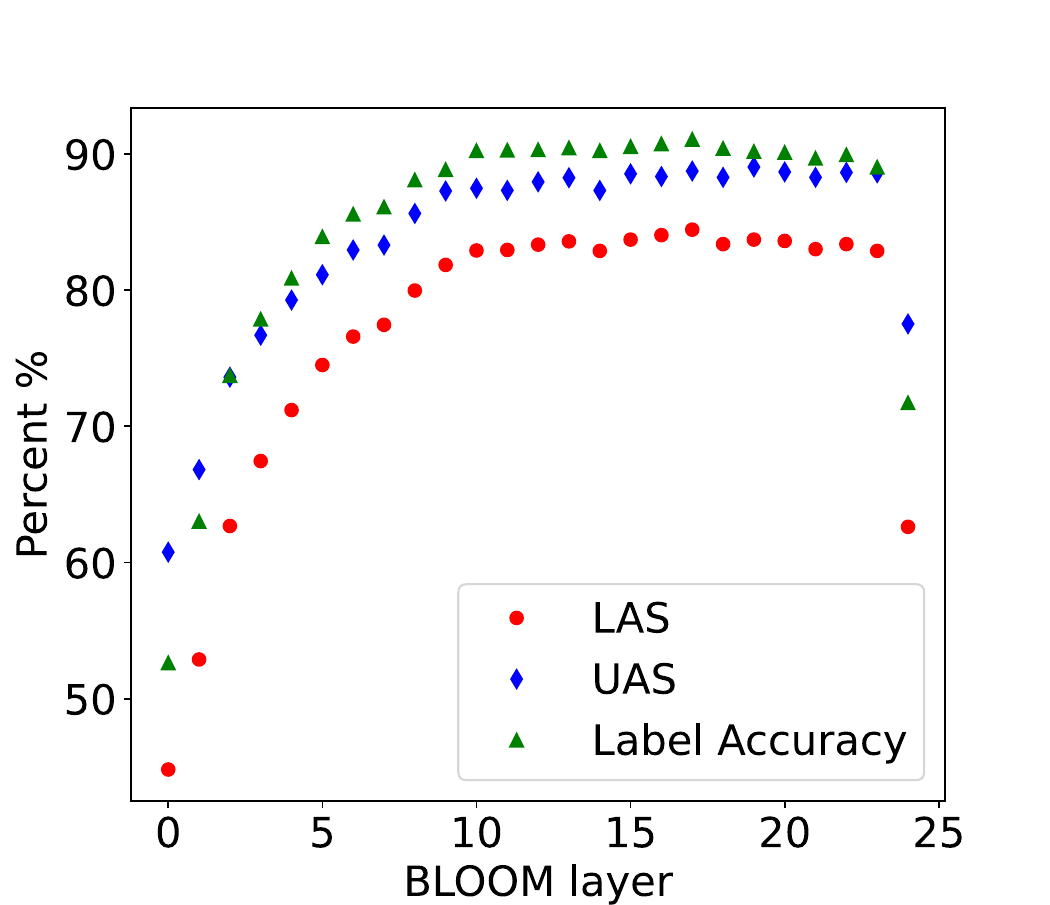}
    \caption{Labeled Accuracy Score (LAS), Unlabeled Accuracy Score (UAS), Label Accuracy for our parser trained and evaluated (dev set) on the Universal Dependencies ParTUT corpus. Each model uses a different BLOOM layer as a sequence encoder.}
    \label{fig:layer_selection}
\end{figure}

We select the 17th layer as the representation to input into the model, which was decided based on optimization of development set accuracy on the ParTUT Universal Dependencies corpus \cite{de2021universal}. We use this smaller corpus (2,090 sentences) in order to conserve resources when investigating the utility of BLOOM representations for dependency parsing. As seen in Figure~\ref{fig:layer_selection}, BLOOM layers 10 to 17 have similar accuracy levels. It is possible that more extensive optimization of this layer~choice
might provide small improvements in accuracy beyond that reported in Table \ref{tab:parser_accuracy}. Further training hyperparameters can be found in the Appendix.

\subsection{Parser Accuracy}
\label{sec:parser_accuracy}

Cognitively~plausible parsers must be incremental, in the sense that they must not use information about as-yet-unprocessed words occurring later in the sentence.
We compare the accuracy of the incremental generative parser proposed here to that of  \citet{buys2018neural}, which employs the same transition system with encodings based on an LSTM with random initial weights. 
    
Parser accuracy is evaluated on the development set after each epoch, and the highest scoring epoch is reported in Table \ref{tab:parser_accuracy}. 
However, the cognitive~modeling results reported in Section \ref{sec:results} rely on earlier training checkpoints. The epoch is chosen by maximizing $r^2$ correlation with the fMRI data. The labeled attachment score and unlabeled attachment score (LAS/UAS) for the epochs chosen are 89.75/92.37 for English, and 77.06/83.59 for Chinese. We find that the objective of obtaining minimal loss does not correspond to optimal correlation with fMRI activity. 

\subsection{Multiple paths} \label{sec:rankedparallel}
Multipath parsing, as a psycholinguistic claim about human sentence comprehension, is simply the idea that ambiguity-resolution pursues more than one alternative at the same time.
This contrasts with Frazier and Fodor's \citeyearpar{frazier78} conception, in which a single parse is developed along one path.
We consider here a version of the multipath idea that \citet{kurtzman-1984-ambiguity} calls ``strong'' parallelism, in which paths can persist from word to word.
\citet{gorrell87} emphasizes that multiple alternatives
are not equally available (page 84). This is naturally formalized by ranking the paths.
Such ranking can be implemented with beam~search, at the heart of
parsing systems like \citet{roark-2001-probabilistic}, which is widely applied in computational psycholinguistics and neurolinguistics.

\subsection{Complexity Metric} \label{sec:complexitymetric}

In order to correlate parser predictions with brain~activity, and in particular to take multiple parser paths into consideration, those predictions must somehow be summarized and quantified.
One way to do that is via the surprisal complexity metric, revived by \citet{hale2001probabilistic} as a method of predicting sentence-processing difficulty word-by-word. The basic~formula is the logarithm of
the reciprocal of a probability.
    \begin{equation}\label{eq:surprisal}
        \mathrm{log}_2\left(\frac{1}{p(y)}\right) = -\mathrm{log}_2(p(y))
    \end{equation}

\noindent In Equation~\ref{eq:surprisal}, $p(y)$ denotes the probability of an arbitrary outcome $y$. In incremental sentence comprehension the relevant outcomes are words. The surprisal of a word at position $w_i$ which follows a string ending in $w_{i-1}$, is a ratio of the probability at the current word with the previous probability:

\begin{equation}\label{eq:incr_surprisal}
    -\mathrm{log}_2\left(\frac{p(w_i, w_{i-1}, ..., w_1))}{p(w_{i-1},..., w_1))}\right)
\end{equation}

\subsubsection{Syntactic Surprisal}

\begin{figure}[t]
    \centering

        \includegraphics[width=\columnwidth]{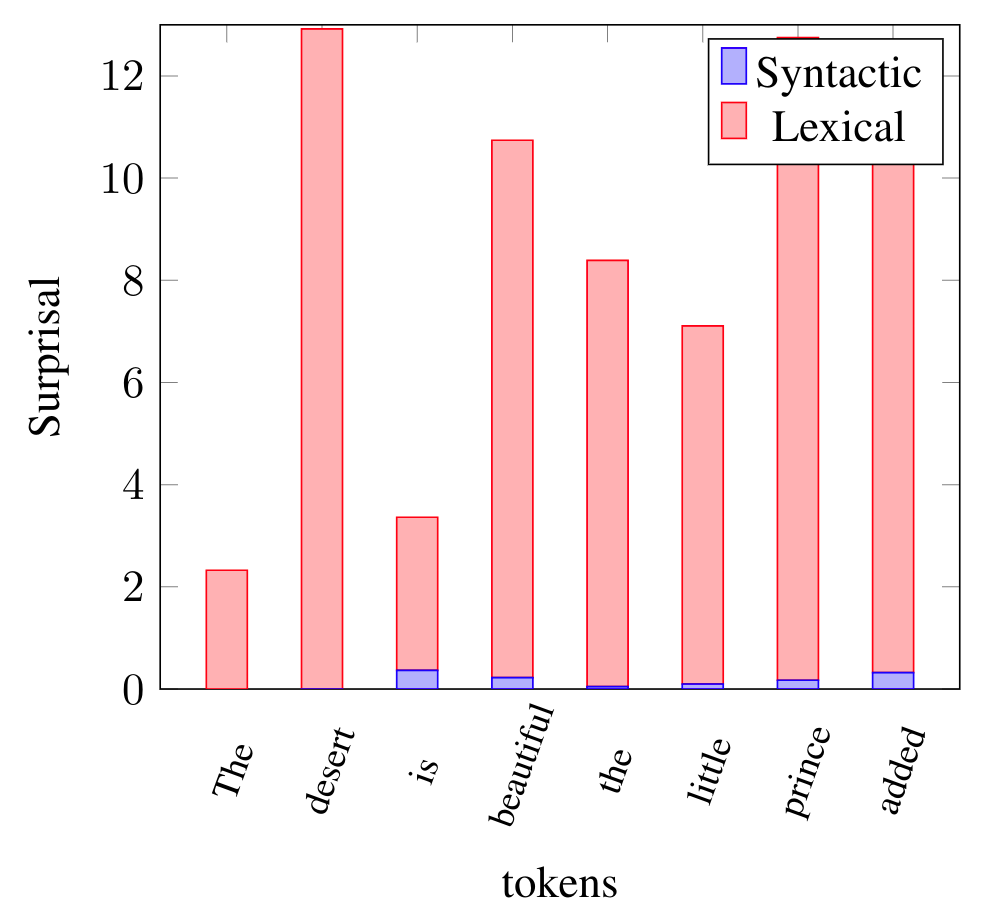}
    \caption{A sentence from \textit{The Little Prince}, showing incremental surprisal from joint probability decomposed into lexical (red), and syntactic surprisal (blue)}
    \label{fig:word_vs_tr}
\end{figure}

Word~probabilities, denoted $p_{gen}$ in Table~\ref{tab:transitions}, are affected by nonsyntactic factors.
As Figure~\ref{fig:word_vs_tr} shows, these nonsyntactic factors can overshadow the difference between syntactic alternatives.
For this reason, we follow \citet{roark-etal-2009-deriving} in decomposing surprisal
into lexical and syntactic expectations.
Discarding lexical~expectations, we focus exclusively on differences among competing syntactic analyses
by adopting their \textit{syntactic surprisal} measure, given below as Equation \ref{eq:synS}.
Syntactic surprisal does not consider word probability at the current time step ($i$), but the complete path probability is used in the denominator ($i-1$) to ensure that the correct path is selected. 

In addition, we limit the sum to the top~$k$ transition sequences. %
Here, $j$~indexes discrete ranks of a ranked parallel parser, and $t_{a(i)}$ refers to all the parser transitions in the path that ends at generation of token $w_i$. At each time step $i$, $t_{a(i)}$ includes 1 shift transition, and zero to many possible reduce transitions.

%% revised to show conditional probability
\begin{align}
\label{eq:synS}  \lefteqn{\mathrm{SynS}_{k}(w_{i}) =} \\ \nonumber
& & -\mathrm{log}_2 
    \frac{\sum_{j=1}^{k} p\left(t_{a(i)} \dots t_{a(1)} \mid
    w_{i-1} \dots\!w_1\right)_{j}}
    {\sum_{j=1}^{k} p\left(t_{a(i-1)} \dots t_{a(1)} \mid w_{i-1}\!\dots\!w_1\right)_j}  
\end{align} 

\begin{figure}[h]
\setlength{\belowcaptionskip}{-8pt}
    \centering
    \includeinkscape[width=\columnwidth]{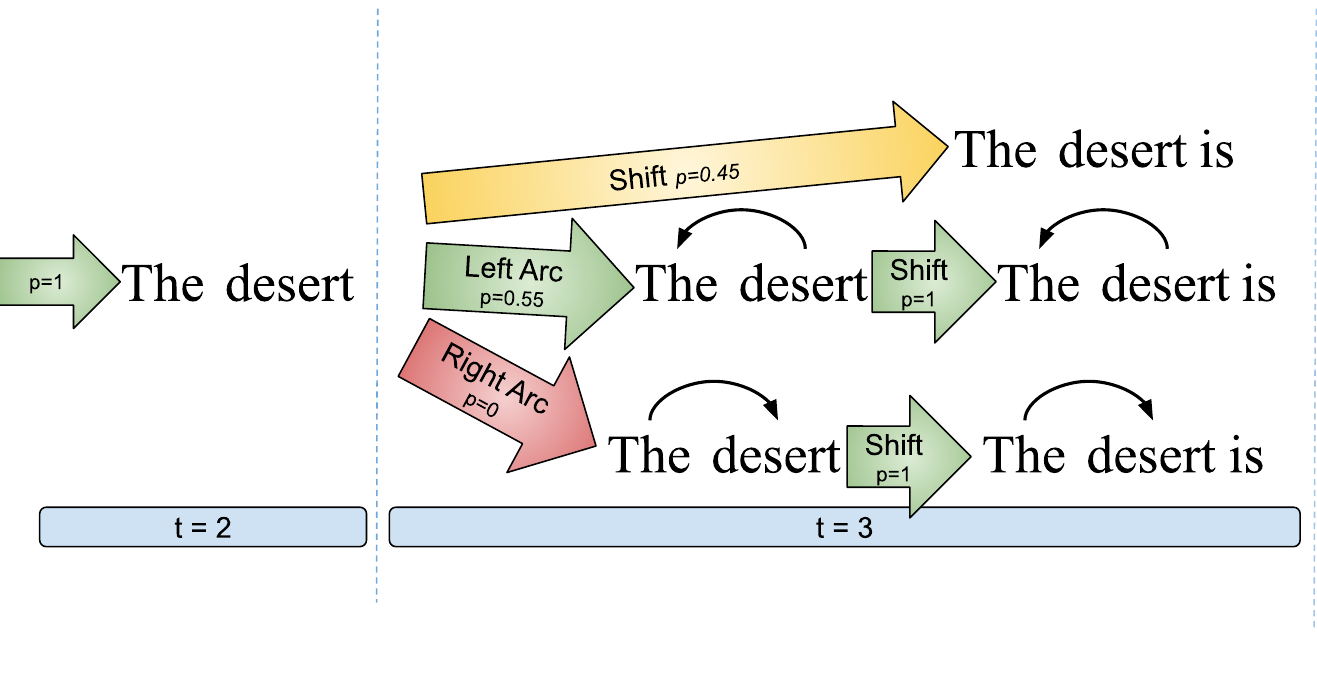}
    \caption{A sentence fragment from \textit{The Little Prince} showing shift-reduce parser actions with associated probabilities. Word generation probabilities are not shown here, but would be included in the full path calculation. In this example, the syntactic surprisal at time t = 3 is -log$_2$(0.55/1) = 0.86 for $k$=1, and for $k$=2 the top two paths are added: -log((0.55+0.45)/1) = 0.}
    \label{fig:parsing_process}
\end{figure}

\noindent In Equation~\ref{eq:synS},  syntactic~surprisal is parameterized by a maximum number of allowable analysis paths, $k$.
Figure~\ref{fig:parsing_process} elaborates an example calculation using this definition.

The limitation to $k$ paths in Equation \ref{eq:synS} follows \citet{boston2011parallel}.
Unlike \citeauthor{boston2011parallel},
we retain all analyses and simply chose the top~$k$ at each successive word. 
This allows paths that would have been lost~forever in true beam~search to later rejoin the beam.
Such restoration has essentially the same effect as backtracking, in cases where it would lead to a higher-scoring analyses.
``Reanalysis'' of this sort could be helpful in modeling the comprehension of garden-path sentences \citep[see e.g.][]{fodor:reanalysis}.

\subsubsection{Why Five Paths?}

\begin{figure*}
    \centering
    \includegraphics[width=0.9\textwidth]{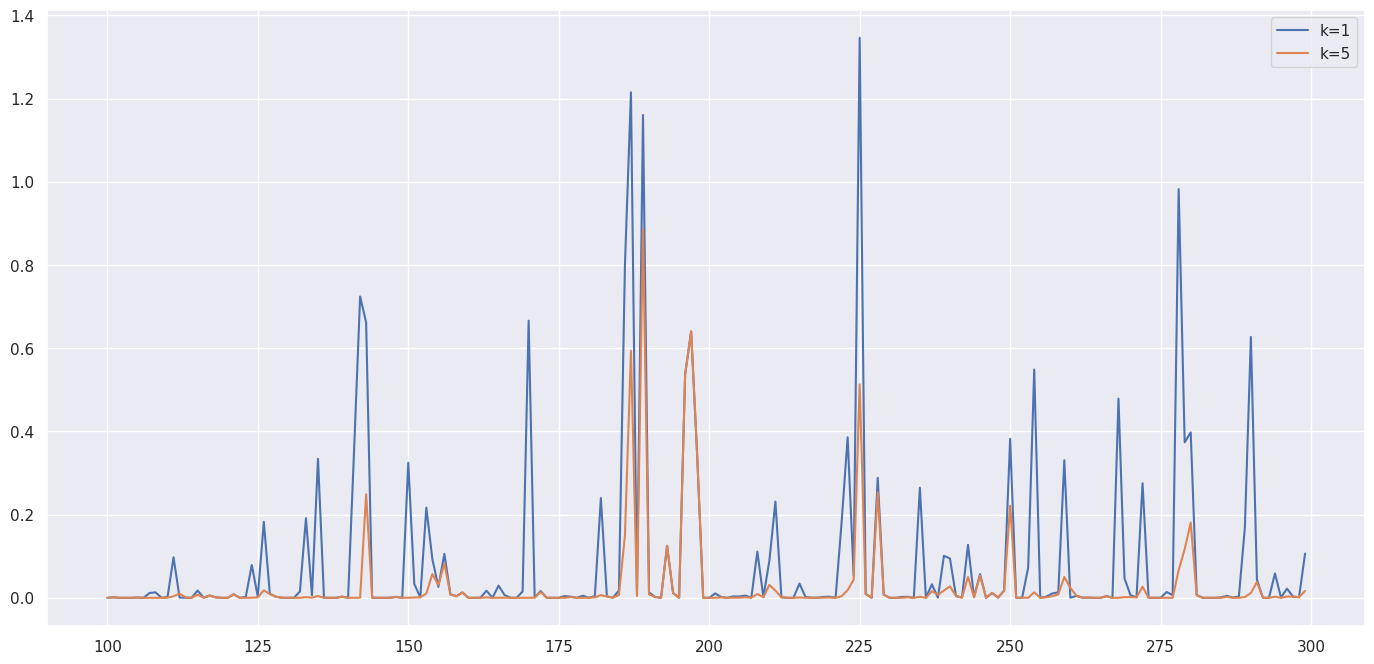}  %
    \caption{Syntactic surprisal at $k=1$ (blue) and $k=5$ (orange) for tokens in \textit{The Little Prince}}
    \label{fig:k1_vs_k5}
\end{figure*}

Addressing the single-path vs multipath question entails choosing some number of paths, $k$ to represent multipath parsing in general. Empirical~literature such as \citet{gibsonpearlmutter00} has been primarily concerned with
the distinction between 1 and 2 paths. By contrast, accurate NLP systems sometimes consider tens or even hundreds of paths.\footnote{Many of these paths
lead to similar syntactic structures. Grouping them in a cognitively realistic way brings~up foundational questions regarding
the mental representation of grammatical relations \citetext{see \citealt{bresnan:competence},
\citealt{sturt96},
and \S4.4 of \citealt{brasoveanu20}, among others}. These questions, along with \label{fn:repair} the proper formulation
of reanalysis or repair operations \citep[e.g.][]{lewis92,mbk01:lops} are important directions for future~work.}
This mismatch calls for a setting of $k$ that is small, but clearly different from 1. The syntactic~surprisal metric itself also mitigates against settings of $k$ that are too large.
With this metric, more ranks quickly eat~up the available probability~mass, up to 1.0, when considering all possible paths.
In this limiting~case syntactic~surprisal would trivially equal zero for all tokens (see Figure~\ref{fig:k1_vs_k5}).  For these reasons, we selected five paths to contrast with single-path parsing.\footnote{This goal of this study is to adjudicate between single-path
and
multipath
parsing. This question has remained open within the literature
on human sentence processing for quite some time.  An alternative~approach seeks to maximize correlations with fMRI data
by e.g. searching for the optimal number of paths. This suggests a natural follow-up.} %

\section{fMRI Methods}

\subsection{Participants}

As detailed in \citet{li2022petit}, the English dataset includes 49 participants (30 female, mean age = 21.3, range = 18-37), and the Chinese dataset includes 35 participants (15 female, mean age = 19.9, range = 18-24). 

\subsection{Data Acquisition}

The English audio stimulus is an English translation of \textit{The Little Prince}, read by Karen~Savage. The Chinese audio stimulus is a Chinese translation of \textit{The Little Prince}, read by a professional female Chinese broadcaster. The English and Chinese audiobooks are 94 and 99 minutes in length, respectively. The presentations were divided into nine sections, each lasting around ten minutes. Participants listened passively to the nine sections and completed four quiz questions after each section (36 questions in total). These questions were used to confirm participant comprehension of the story.
    
The English and Chinese brain imaging data were acquired with a 3T MRI GE Discovery MR750 scanner with a 32-channel head coil. Anatomical scans were acquired using a T1-weighted volumetric magnetization prepared rapid gradient-echo pulse sequence. Blood-oxygen-level-dependent (BOLD) functional scans were acquired using a multi-echo planar imaging sequence with online reconstruction (TR = 2000 ms; TE’s = 12.8, 27.5, 43 ms; FA = 77\textdegree; matrix size = 72 x 72; FOV = 240.0 mm x 240.0 mm; 2x image acceleration; 33 axial slices, voxel size = 3.75 x 3.75 x 3.8 mm). 
    
\subsection{Data Preprocessing}

The English and Chinese fMRI data were preprocessed using AFNI version 16 \citep{cox1996afni}. The first 4 volumes in each run were excluded from analyses to allow for T1-equilibration effects. Multi-echo independent components analysis (ME-ICA), was used to denoise data for motion, physiology, and scanner artifacts \cite{kundu2012differentiating}. Images were then spatially normalized to the standard space of the Montreal Neurological Institute (MNI) atlas, yielding a volumetric time series resampled at 2 mm cubic voxels.

\subsection{Statistical Analysis}\label{sec:Statistical}
The goal of the analysis is to compare
surprisal~values from single-path and multipath parsers against the same observed fMRI timecourses.
We follow \citet{crabbe-etal-2019-variable} in evaluating goodness~of~fit pairwise using cross-validated coefficient of determination ($r^2$) maps.

For each subject individually, the fMRI BOLD~signal is modeled by a General Linear Model (GLM) at each voxel. The following five regressors are included in the GLM:  word~rate, fundamental~frequency (f0), word~frequency, root mean~square intensity (RMS), and syntactic~surprisal of the top $k$~paths. 
Word rate is a timing function marking the offset of each spoken word; f0 is the fundamental frequency, or pitch of the audio; word frequency is obtained from words in a movie subtitles database \cite{brysbaert2009moving}; and RMS is an indicator of audio intensity, taken every 10ms.
These predictors are known to affect speech comprehension, and are included as control~variables to help isolate variance within the BOLD~signal that is specific to syntactic~processing.
    
Syntactic surprisal for each word was also aligned to the offset of each word in the audiobook. All predictors were convolved using SPM’s canonical Hemodynamic Response Function \cite{friston2007statistical}.

For each participant, we compute how much the inclusion of the syntactic~surprisal regressor increases the cross-validated $r^2$ %
over the baseline model, which includes only control~variables. We repeat this process for syntactic~surprisal at different levels of $k$ separately in order to keep the number of parameters in each GLM model constant. Therefore, the $r^2$ \textit{increase}  scores represent the variance explained in each voxel by the addition of syntactic~surprisal to the model as a predictor.

To compare models across two different levels of $k$ and analyze their ability to explain the fMRI BOLD signal, we performed a paired t-test on individual $r^2$ \textit{increase} maps to obtain z-maps.
These z-maps show where syntactic~surprisal from a model with a given number of paths explains the signal significantly~better than does a model that incorporates a different number of paths (see Figures \ref{fig:english_nongen} and \ref{fig:chinese_nongen}).
    
\section{Results} \label{sec:results}
\begin{figure*}
    \centering
    \includegraphics[width=0.75\textwidth]{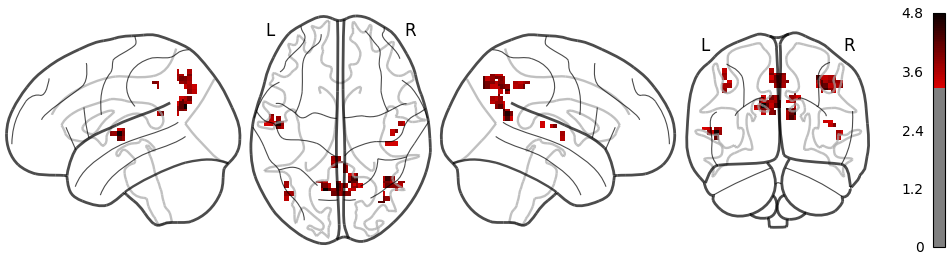}
    \caption{English Brain z-maps showing the significant clusters (p < .001 uncorrected; cluster threshold = 15 voxels) for the model comparison between syntactic surprisal at $k=1$ vs $k=5$. All significant clusters show greater $r^2$ increase for surprisal at $k=5$ than for surprisal at $k=1$. }
    \label{fig:english_nongen}    
\end{figure*}
\begin{figure*}
    \centering
    \includegraphics[width=0.75\textwidth]{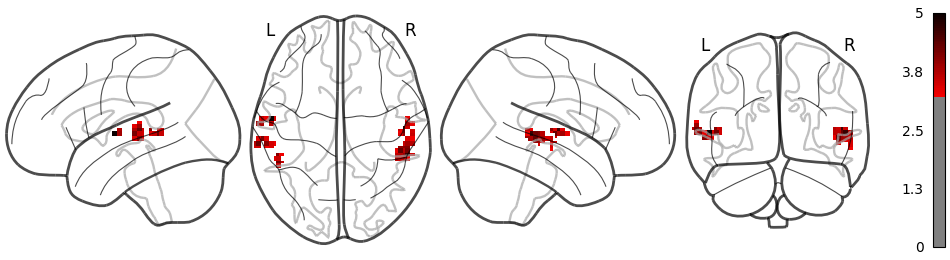}
    \caption{Chinese Brain z-maps showing the significant clusters (p < .001 uncorrected; cluster threshold = 15 voxels) for the model comparison between syntactic surprisal at $k=1$ vs $k=5$. All significant clusters show greater $r^2$ increase for surprisal at $k=5$ than for surprisal at $k=1$.}
    \label{fig:chinese_nongen}
\end{figure*}
Both English and Chinese paired t-tests on $r^2$ increase maps show differences in the superior temporal gyrus (STG). 
The Chinese results further show differences in middle temporal gyrus (MTG), while the English results show differences in the parietal~lobe,
including bilateral angular gyrus. The full~list of statistically significant clusters corresponding to Figures \ref{fig:english_nongen} and \ref{fig:chinese_nongen} can be found in the Appendix. Although the paired t-tests were two-tailed, all of the resulting significant clusters show greater $r^2$ increase for syntactic~surprisal at $k=5$ than at $k=1$. The $r^2$ increase maps for each model individually can be found in the Appendix.

\section{Discussion} \label{sec:discussion}
Multipath parsing effort, outside of the 1-best analysis, localizes to superior temporal regions of the brain bilaterally in both English and Chinese. 
This finding converges with \citet{crabbe-etal-2019-variable}. Crabb\'{e} and colleagues parse phrase~structure, rather than dependencies,
employ a different complexity~metric,
and use a very~large beam (400).
Despite all these differences, their results implicate roughly the same brain~regions. 

These results also cohere with proposals regarding the large-scale organization of language processing in the brain. On the Neuroanatomical Pathway model,
for instance, ``most basic syntactic processes'' are handled by a ventral~pathway that passes through STG regions identified here \citep{friederici:pathway}.
\citet{matchin2020cortical} similarly point to the ventral~pathway as essential for ``basic'' syntactic processes.
This heuristic~notion of basic syntactic processing aligns well with Stanford~Dependencies' avowed goal of providing a simple, surface-oriented notation for local grammatical relations.

The present results do not identify a significant difference in $r^2$ increase in Broca's area (left IFG).
Some increase in $r^2$ can be seen, however, in middle and IFG in the individual model maps shown in the Appendix. 
The absence of a significant difference, however makes~sense in~light~of two considerations. The first is the idea that left IFG subserves reanalysis in the face of misinterpretation \citep{novick-cognitive-2005}.
The second is the literary style of the \textit{The Little Prince}.
The stimulus~text is dissimilar to the sorts of garden-path materials psycholinguists have used \citetext{see e.g. \citealt[][chapter~9]{sedivy19} or \citealt[][chapter~7]{fernandez10}}.
It seems likely that any garden-pathing was mild, remaining below the level of conscious awareness.
This distinction between misinterpretations that rise to the level of conscious awareness and those that do~not
could help reconcile these results with earlier studies showing IFG activation in response to garden path stimuli \citep[e.g.][]{mason2003ambiguity,rodd2010functional}.
For instance, \citet{jager2015subject} suggests that Chinese relative~clauses involve as many as four temporary ambiguities. To our knowledge none of these reach the level of conscious awareness in everyday comprehension.
Yet in a naturalistic fMRI study, \citet{dunagan2022neural} observe activation in anterior and middle STG that is specific to Chinese object-extracted relative~clauses.
This contrast was not reliable in a statistical comparison between English and Chinese.
Further work is needed to chart the space between unproblematic~ambiguities \citep[][\S 2.4.2]{lewis93} and conscious misinterpretations engendered by syntactic ambiguity.
These STG results include activation in the primary auditory cortex, which could suggest, in line with the sensory hypothesis \citep{dikker2009sensitivity}, that expectations based on earlier syntactic processing first impact regions involved in low-level sensory processing.
This possibility could be examined in a followup fMRI study with written or signed materials, along the lines of \citet{Henderson:2016yq}.

\section{Conclusion and Future Work}
The main conclusion is that human parsing is multipath.
This follows from observing greater $r^{2}$ increase for multipath surprisal than single-path surprisal in fMRI data.
Even with a very bottom-up strategy, such as the arc-hybrid system used here, it seems that more than one path must be pursued in order to best-align with humans' word-by-word effort profile.
This conclusion is consistent with
disjunctive~representations of choices such as PP~attachment \citep[][\S5.5]{kitaev-etal-2022-learned}.

The multipath interpretation that we offer here can be confirmed or refuted with other linked linguistic and neuro-cognitive databases -- for instance,  in different genres or languages. In addition, using brain data with a higher temporal resolution, such as MEG, may provide a benefit given the temporary nature of the syntactic ambiguities included in our model.   
Although we take the primary finding to be one of commonality across STG regions in both languages, English listeners did uniquely show an additional~effect of multipath parsing in parietal~regions.
Activation in this area has been correlated with measures of human memory such as digit~span \citep{meyer2012linking}. Individuals' memory~span may modulate the number of paths that they pursue during comprehension \citep{vos01,prat10}. Future~work should address the interaction between disambiguation and memory \citep[e.g.][]{campanelli-modulatory-2018}.

\section*{Limitations}
The goal of correlating parser~states with fMRI~data is limited by the particularities of the parsing system -- namely the arc-hybrid transition system and the limited size and content of the training data (which is dissimilar in genre to the audiobook text we study). 
In addition, the English results in particular are the result of careful model selection to limit ``training away'' the syntactic ambiguity we would like to measure. This may indicate that more formalized external limitations should be applied to modern high-performing parsers should they be used to model human sentence processing. 

\section*{Ethics}
Language models pose risks when used outside of their intended scope. We use BLOOM, which is publicly available under the Responsible AI License (RAIL) \cite{scao2022bloom}. Our scientific enquiry falls within the intended use of public research on LLMs. We also use a publicly available fMRI dataset \cite{li2022petit}, which is available under a CCO license. This dataset was anonymized to remove identifying facial features before publication.

\section*{Acknowledgements}
 This material is based upon work supported by the National Science Foundation under Grant Number 1903783.
 We thank Laura Rimell and Michael Covington for valuable discussion, along with Christophe Pallier for developing the cross-validated $r^{2}$ code used here.
 We are grateful to many collaborators for assistance with the fMRI data collection.
Among them are Shohini Bhattasali, Jixing Li, Wen-Ming Luh and Nathan Spreng.

\bibliography{anthology,custom}

\begin{thebibliography}{70}
\expandafter\ifx\csname natexlab\endcsname\relax\def\natexlab#1{#1}\fi

\bibitem[{Bornkessel-Schlesewsky and Schlesewsky(2015)}]{adm:30}
Ina Bornkessel-Schlesewsky and Matthias Schlesewsky. 2015.
\newblock \href {https://doi.org/10.1016/C2011-0-07351-9} {The {A}rgument
  {D}ependency {M}odel}.
\newblock In  \cite{hickoksmall15}, chapter~30.

\bibitem[{Boston et~al.(2011)Boston, Hale, Vasishth, and
  Kliegl}]{boston2011parallel}
Marisa~Ferrara Boston, John~T Hale, Shravan Vasishth, and Reinhold Kliegl.
  2011.
\newblock Parallel processing and sentence comprehension difficulty.
\newblock \emph{Language and Cognitive Processes}, 26(3):301--349.

\bibitem[{Brasoveanu and Dotla\v{c}il(2020)}]{brasoveanu20}
Adrian Brasoveanu and Jakub Dotla\v{c}il. 2020.
\newblock \emph{Computational Cognitive Modeling and Linguistic Theory}.
\newblock Language, Cognition, and Mind. Springer.

\bibitem[{Brennan et~al.(2020)Brennan, Dyer, Kuncoro, and
  Hale}]{brennan:2020ku}
Jonathan~R. Brennan, Chris Dyer, Adhiguna Kuncoro, and John~T. Hale. 2020.
\newblock \href {https://doi.org/10/ghd38w} {Localizing syntactic predictions
  using recurrent neural network grammars}.
\newblock \emph{Neuropsychologia}, 146:1074--1079.

\bibitem[{Brennan et~al.(2016)Brennan, Stabler, Van~Wagenen, Luh, and
  Hale}]{brennan2016-tv}
Jonathan~R. Brennan, Edward~P. Stabler, Sarah~E Van~Wagenen, Wen-Ming Luh, and
  John~T. Hale. 2016.
\newblock Abstract linguistic structure correlates with temporal activity
  during naturalistic comprehension.
\newblock \emph{Brain Lang.}, 157-158:81--94.

\bibitem[{Bresnan and Kaplan(1982)}]{bresnan:competence}
Joan Bresnan and Ronald~M. Kaplan. 1982.
\newblock Introduction: Grammars as mental representations of language.
\newblock In Joan Bresnan, editor, \emph{The Mental Representation of
  Grammatical Relations}, pages xvii,lii. MIT Press, Cambridge, MA.

\bibitem[{Brysbaert and New(2009)}]{brysbaert2009moving}
Marc Brysbaert and Boris New. 2009.
\newblock Moving beyond {K}u{\v{c}}era and {F}rancis: A critical evaluation of
  current word frequency norms and the introduction of a new and improved word
  frequency measure for {A}merican {E}nglish.
\newblock \emph{Behavior research methods}, 41(4):977--990.

\bibitem[{{Buch-Kromann}(2004)}]{mbk01:lops}
Matthias {Buch-Kromann}. 2004.
\newblock \href {https://doi.org/10.1016/S1571-0661(05)82581-2} {Optimality
  parsing and local cost functions in {D}iscontinuous {G}rammar}.
\newblock \emph{Electronic Notes in Theoretical Computer Science}, 53:163--179.
\newblock Proceedings of the joint meeting of the 6th Conference on Formal
  Grammar and the 7th Conference on Mathematics of Language.

\bibitem[{Buys and Blunsom(2018)}]{buys2018neural}
Jan Buys and Phil Blunsom. 2018.
\newblock Neural syntactic generative models with exact marginalization.
\newblock In \emph{Proceedings of the 2018 Conference of the North American
  Chapter of the Association for Computational Linguistics: Human Language
  Technologies, Volume 1 (Long Papers)}, pages 942--952.

\bibitem[{Campanelli et~al.(2018)Campanelli, Van~Dyke, and
  Marton}]{campanelli-modulatory-2018}
Luca Campanelli, Julie~A. Van~Dyke, and Klara Marton. 2018.
\newblock The modulatory effect of expectations on memory retrieval during
  sentence comprehension.
\newblock In Timothy~T. Rogers, Marina Rau, Xiaojin Zhu, and Charles~W. Kalish,
  editors, \emph{Proceedings of the 40th {Annual} {Conference} of the
  {Cognitive} {Science} {Society}}, pages 1434--1439. Cognitive Science
  Society, Austin, Texas.

\bibitem[{Caucheteux and King(2022)}]{caucheteux22}
Charlotte Caucheteux and Jean-R{\'e}mi King. 2022.
\newblock \href {https://doi.org/10.1038/s42003-022-03036-1} {Brains and
  algorithms partially converge in natural language processing}.
\newblock \emph{Nature Communications Biology}, 5(1):134.

\bibitem[{Cox(1996)}]{cox1996afni}
Robert~W Cox. 1996.
\newblock Afni: software for analysis and visualization of functional magnetic
  resonance neuroimages.
\newblock \emph{Computers and Biomedical research}, 29(3):162--173.

\bibitem[{Crabb{\'e} et~al.(2019)Crabb{\'e}, Fabre, and
  Pallier}]{crabbe-etal-2019-variable}
Benoit Crabb{\'e}, Murielle Fabre, and Christophe Pallier. 2019.
\newblock \href {https://doi.org/10.18653/v1/D19-1106} {Variable beam search
  for generative neural parsing and its relevance for the analysis of
  neuro-imaging signal}.
\newblock In \emph{Proceedings of the 2019 Conference on Empirical Methods in
  Natural Language Processing and the 9th International Joint Conference on
  Natural Language Processing (EMNLP-IJCNLP)}, pages 1150--1160, Hong Kong,
  China. Association for Computational Linguistics.

\bibitem[{De~Marneffe and Manning(2008)}]{de2008stanford}
Marie-Catherine De~Marneffe and Christopher~D Manning. 2008.
\newblock The stanford typed dependencies representation.
\newblock In \emph{Coling 2008: proceedings of the workshop on cross-framework
  and cross-domain parser evaluation}, pages 1--8.

\bibitem[{De~Marneffe et~al.(2021)De~Marneffe, Manning, Nivre, and
  Zeman}]{de2021universal}
Marie-Catherine De~Marneffe, Christopher~D Manning, Joakim Nivre, and Daniel
  Zeman. 2021.
\newblock Universal dependencies.
\newblock \emph{Computational linguistics}, 47(2):255--308.

\bibitem[{Dikker et~al.(2009)Dikker, Rabagliati, and
  Pylkk{\"a}nen}]{dikker2009sensitivity}
Suzanne Dikker, Hugh Rabagliati, and Liina Pylkk{\"a}nen. 2009.
\newblock Sensitivity to syntax in visual cortex.
\newblock \emph{Cognition}, 110(3):293--321.

\bibitem[{Dunagan et~al.(2022)Dunagan, Stanojevi{\'c}, Coavoux, Zhang,
  Bhattasali, Li, Brennan, and Hale}]{dunagan2022neural}
Donald Dunagan, Milo{\v{s}} Stanojevi{\'c}, Maximin Coavoux, Shulin Zhang,
  Shohini Bhattasali, Jixing Li, Jonathan Brennan, and John Hale. 2022.
\newblock Neural correlates of object-extracted relative clause processing
  across {E}nglish and {C}hinese.
\newblock \emph{Neurobiology of Language}, pages 1--43.

\bibitem[{Eisape et~al.(2022)Eisape, Gangireddy, Levy, and
  Kim}]{eisape-etal-2022-probing}
Tiwalayo Eisape, Vineet Gangireddy, Roger Levy, and Yoon Kim. 2022.
\newblock \href {https://doi.org/10.18653/v1/2022.findings-emnlp.203} {Probing
  for incremental parse states in autoregressive language models}.
\newblock In \emph{Findings of the Association for Computational Linguistics:
  EMNLP 2022}, pages 2801--2813, Abu Dhabi, United Arab Emirates. Association
  for Computational Linguistics.

\bibitem[{Fern{\'{a}}ndez and Cairns(2010)}]{fernandez10}
Eva~M. Fern{\'{a}}ndez and Helen~Smith Cairns. 2010.
\newblock \emph{Fundamentals of Psycholinguistics}.
\newblock Wiley-Blackwell.

\bibitem[{Fodor and Ferreira(1998)}]{fodor:reanalysis}
Janet~Dean Fodor and Fernanda Ferreira, editors. 1998.
\newblock \emph{Reanalysis in sentence processing}, volume~21 of \emph{Studies
  in Theoretical Psycholinguistics}.
\newblock Kluwer, Dordrecht.

\bibitem[{Frazier and Fodor(1978)}]{frazier78}
Lyn Frazier and Janet~Dean Fodor. 1978.
\newblock The sausage machine: a new two-stage parsing model.
\newblock \emph{Cognition}, 6:291--325.

\bibitem[{Friederici(2015)}]{friederici:pathway}
Angela~D. Friederici. 2015.
\newblock \href {https://doi.org/10.1016/B978-0-12-407794-2.00029-8} {The
  neuroanatomical pathway model of language}.
\newblock In  \cite{hickoksmall15}, chapter~29.

\bibitem[{Friston et~al.(2007)Friston, Ashburner, Kiebel, Nichols, and
  Penny}]{friston2007statistical}
Karl~J Friston, John Ashburner, Stefan~J Kiebel, T~Nichols, and William Penny.
  2007.
\newblock \emph{Statistical Parametric Mapping}.
\newblock Academic Press.

\bibitem[{Gibson(1991)}]{gibson91}
Edward Gibson. 1991.
\newblock \emph{A Computational Theory of Human Linguistic Processing: Memory
  Limitations and Processing Breakdown}.
\newblock Ph.D. thesis, Carnegie Mellon University.

\bibitem[{Gibson and Pearlmutter(2000)}]{gibsonpearlmutter00}
Edward Gibson and Neal~J. Pearlmutter. 2000.
\newblock Distinguishing serial and parallel parsing.
\newblock \emph{Journal of Psycholinguistic Research}, 29(2).

\bibitem[{Gorrell(1987)}]{gorrell87}
Paul~G. Gorrell. 1987.
\newblock \href {https://digitalcommons.lib.uconn.edu/dissertations/AAI8728869}
  {\emph{Studies of human syntactic processing: Ranked-parallel versus serial
  models}}.
\newblock Ph.D. thesis, University of {C}onnecticut.

\bibitem[{Hale(2016)}]{hale2016-dm}
John Hale. 2016.
\newblock Information-theoretical complexity metrics.
\newblock \emph{Language and Linguistics Compass}, 10(9):397--412.

\bibitem[{Hale et~al.(2018)Hale, Dyer, Kuncoro, and
  Brennan}]{hale-etal-2018-finding}
John Hale, Chris Dyer, Adhiguna Kuncoro, and Jonathan Brennan. 2018.
\newblock \href {https://doi.org/10.18653/v1/P18-1254} {Finding syntax in human
  encephalography with beam search}.
\newblock In \emph{Proceedings of the 56th Annual Meeting of the Association
  for Computational Linguistics (Volume 1: Long Papers)}, pages 2727--2736,
  Melbourne, Australia. Association for Computational Linguistics.

\bibitem[{Hale(2001)}]{hale2001probabilistic}
John~T Hale. 2001.
\newblock A probabilistic {E}arley parser as a psycholinguistic model.
\newblock In \emph{Second meeting of the {N}orth {A}merican chapter of the
  association for computational linguistics}.

\bibitem[{Hale et~al.(2022)Hale, Campanelli, Li, Bhattasali, Pallier, and
  Brennan}]{hale2022neurocomputational}
John~T Hale, Luca Campanelli, Jixing Li, Shohini Bhattasali, Christophe
  Pallier, and Jonathan~R Brennan. 2022.
\newblock Neurocomputational models of language processing.
\newblock \emph{Annual Review of Linguistics}, 8:427--446.

\bibitem[{Henderson et~al.(2016)Henderson, Choi, Lowder, and
  Ferreira}]{Henderson:2016yq}
John~M Henderson, Wonil Choi, Matthew~W Lowder, and Fernanda Ferreira. 2016.
\newblock \href {https://doi.org/10.1016/j.neuroimage.2016.02.050} {Language
  structure in the brain: {{A}} fixation-related {{fMRI}} study of syntactic
  surprisal in reading}.
\newblock \emph{NeuroImage}, 132:293--300.

\bibitem[{Hickok et~al.(2015)Hickok, Small, and Small}]{hickoksmall15}
Gregory Hickok, Steven~L. Small, and Steven~L. Small. 2015.
\newblock \href {https://doi.org/10.1016/C2011-0-07351-9} {\emph{Neurobiology
  of Language}}.
\newblock Elsevier, San Diego.

\bibitem[{Hopf et~al.(2003)Hopf, Bader, Meng, and Bayer}]{hopf03}
Jens-Max Hopf, Markus Bader, Michael Meng, and Josef Bayer. 2003.
\newblock \href {https://doi.org/https://doi.org/10.1016/S0926-6410(02)00149-0}
  {Is human sentence parsing serial or parallel?: Evidence from event-related
  brain potentials}.
\newblock \emph{Cognitive Brain Research}, 15(2):165--177.

\bibitem[{J{\"a}ger et~al.(2015)J{\"a}ger, Chen, Li, Lin, and
  Vasishth}]{jager2015subject}
Lena J{\"a}ger, Zhong Chen, Qiang Li, Chien-Jer~Charles Lin, and Shravan
  Vasishth. 2015.
\newblock The subject-relative advantage in {C}hinese: Evidence for
  expectation-based processing.
\newblock \emph{Journal of Memory and Language}, 79:97--120.

\bibitem[{Jurafsky(1996)}]{jurafsky1996probabilistic}
Daniel Jurafsky. 1996.
\newblock A probabilistic model of lexical and syntactic access and
  disambiguation.
\newblock \emph{Cognitive science}, 20(2):137--194.

\bibitem[{Kitaev et~al.(2022)Kitaev, Lu, and Klein}]{kitaev-etal-2022-learned}
Nikita Kitaev, Thomas Lu, and Dan Klein. 2022.
\newblock \href {https://doi.org/10.18653/v1/2022.acl-long.220} {Learned
  incremental representations for parsing}.
\newblock In \emph{Proceedings of the 60th Annual Meeting of the Association
  for Computational Linguistics (Volume 1: Long Papers)}, pages 3086--3095,
  Dublin, Ireland. Association for Computational Linguistics.

\bibitem[{Kuhlmann et~al.(2011)Kuhlmann, G{\'o}mez-Rodr{\'\i}guez, and
  Satta}]{kuhlmann2011dynamic}
Marco Kuhlmann, Carlos G{\'o}mez-Rodr{\'\i}guez, and Giorgio Satta. 2011.
\newblock \href {https://aclanthology.org/P11-1068} {Dynamic programming
  algorithms for transition -based dependency parsers}.
\newblock In \emph{Proceedings of the 49th Annual Meeting of the Association
  for Computational Linguistics: Human Language Technologies}, pages 673--682,
  Portland, Oregon, USA. Association for Computational Linguistics.

\bibitem[{Kundu et~al.(2012)Kundu, Inati, Evans, Luh, and
  Bandettini}]{kundu2012differentiating}
Prantik Kundu, Souheil~J Inati, Jennifer~W Evans, Wen-Ming Luh, and Peter~A
  Bandettini. 2012.
\newblock Differentiating {BOLD} and non-{BOLD} signals in {fMRI} time series
  using multi-echo {EPI}.
\newblock \emph{Neuroimage}, 60(3):1759--1770.

\bibitem[{Kurtzman(1984)}]{kurtzman-1984-ambiguity}
Howard~S. Kurtzman. 1984.
\newblock \href {https://doi.org/10.3115/980491.980594} {Ambiguity resolution
  in the human syntactic parser: An experimental study}.
\newblock In \emph{10th International Conference on Computational Linguistics
  and 22nd Annual Meeting of the Association for Computational Linguistics},
  pages 481--485, Stanford, California, USA. Association for Computational
  Linguistics.

\bibitem[{Le~Scao et~al.(2022{\natexlab{a}})Le~Scao, Fan, Akiki, Pavlick,
  Ili{\'c}, Hesslow, Castagn{\'e}, Luccioni, Yvon, Gall{\'e}
  et~al.}]{scao2022bloom}
Teven Le~Scao, Angela Fan, Christopher Akiki, Ellie Pavlick, Suzana Ili{\'c},
  Daniel Hesslow, Roman Castagn{\'e}, Alexandra~Sasha Luccioni, Fran{\c{c}}ois
  Yvon, Matthias Gall{\'e}, et~al. 2022{\natexlab{a}}.
\newblock Bloom: A 176b-parameter open-access multilingual language model.
\newblock \emph{arXiv preprint 2211.05100}.

\bibitem[{Le~Scao et~al.(2022{\natexlab{b}})Le~Scao, Wang, Hesslow, Bekman,
  Bari, Biderman, Elsahar, Muennighoff, Phang, Press, Raffel, Sanh, Shen,
  Sutawika, Tae, Yong, Launay, and Beltagy}]{le-scao-etal-2022-language}
Teven Le~Scao, Thomas Wang, Daniel Hesslow, Stas Bekman, M~Saiful Bari, Stella
  Biderman, Hady Elsahar, Niklas Muennighoff, Jason Phang, Ofir Press, Colin
  Raffel, Victor Sanh, Sheng Shen, Lintang Sutawika, Jaesung Tae, Zheng~Xin
  Yong, Julien Launay, and Iz~Beltagy. 2022{\natexlab{b}}.
\newblock \href {https://doi.org/10.18653/v1/2022.findings-emnlp.54} {What
  language model to train if you have one million {GPU} hours?}
\newblock In \emph{Findings of the Association for Computational Linguistics:
  EMNLP 2022}, pages 765--782, Abu Dhabi, United Arab Emirates. Association for
  Computational Linguistics.

\bibitem[{Lewis(1992)}]{lewis92}
Richard~L. Lewis. 1992.
\newblock Recent {D}evelopments in the {NL-Soar} {G}arden {P}ath {T}heory.
\newblock Technical Report CMU-CS-92-141, Carnegie Mellon University.

\bibitem[{Lewis(1993)}]{lewis93}
Richard~L. Lewis. 1993.
\newblock \emph{An {A}rchitecturally-based {T}heory of {H}uman {S}entence
  {C}omprehension}.
\newblock Ph.D. thesis, Carnegie Mellon University, Pittsburgh, {PA}.

\bibitem[{Lewis(2000{\natexlab{a}})}]{lewis00}
Richard~L. Lewis. 2000{\natexlab{a}}.
\newblock Falsifying serial and parallel parsing models: empirical conundrums
  and an overlooked paradigm.
\newblock \emph{Journal of Psycholinguistic Research}, 29(2).

\bibitem[{Lewis(2000{\natexlab{b}})}]{lewis:arch}
Richard~L. Lewis. 2000{\natexlab{b}}.
\newblock Specifying architectures for language processing: Process, control,
  and memory in parsing and interpretation.
\newblock In Matthew~W. Crocker, Martin Pickering, and Charles {Clifton, Jr.},
  editors, \emph{Architectures and mechanisms for language processing}.
  Cambridge University Press.

\bibitem[{Li et~al.(2022)Li, Bhattasali, Zhang, Franzluebbers, Luh, Spreng,
  Brennan, Yang, Pallier, and Hale}]{li2022petit}
Jixing Li, Shohini Bhattasali, Shulin Zhang, Berta Franzluebbers, Wen-Ming Luh,
  R~Nathan Spreng, Jonathan~R Brennan, Yiming Yang, Christophe Pallier, and
  John Hale. 2022.
\newblock Le {P}etit {P}rince multilingual naturalistic {fMRI} corpus.
\newblock \emph{Scientific data}, 9(1):530.

\bibitem[{Li and Hale(2019)}]{jixing:minparse}
Jixing Li and John Hale. 2019.
\newblock Grammatical predictors for {fMRI} timecourses.
\newblock In Robert~C. Berwick and Edward~P. Stabler, editors, \emph{Minimalist
  Parsing}. Oxford University Press.

\bibitem[{Lopopolo et~al.(2021)Lopopolo, van~den Bosch, Petersson, and
  Willems}]{lopopolo21}
Alessandro Lopopolo, Antal van~den Bosch, Karl-Magnus Petersson, and Roel~M.
  Willems. 2021.
\newblock \href {https://doi.org/10.1162/nol_a_00029} {Distinguishing syntactic
  operations in the brain: Dependency and phrase-structure parsing}.
\newblock \emph{Neurobiology of Language}, 2(1):152--175.

\bibitem[{Marcus(1980)}]{marcus80}
Mitchell~P. Marcus. 1980.
\newblock \emph{A theory of syntactic recognition for natural language}.
\newblock MIT Press.

\bibitem[{Marr(1982)}]{marr:vision}
David Marr. 1982.
\newblock \emph{Vision: A computational investigation into the human
  representation and processing of visual information}.
\newblock W.H. Freeman and Company.

\bibitem[{Mason et~al.(2003)Mason, Just, Keller, and
  Carpenter}]{mason2003ambiguity}
Robert~A Mason, Marcel~Adam Just, Timothy~A Keller, and Patricia~A Carpenter.
  2003.
\newblock Ambiguity in the brain: what brain imaging reveals about the
  processing of syntactically ambiguous sentences.
\newblock \emph{Journal of Experimental Psychology: Learning, Memory, and
  Cognition}, 29(6):1319.

\bibitem[{Matchin and Hickok(2020)}]{matchin2020cortical}
William Matchin and Gregory Hickok. 2020.
\newblock \href {https://doi.org/10.1093/cercor/bhz180} {The cortical
  organization of syntax}.
\newblock \emph{Cerebral Cortex}, 30(3):1481--1498.

\bibitem[{Meyer et~al.(2012)Meyer, Obleser, Anwander, and
  Friederici}]{meyer2012linking}
Lars Meyer, Jonas Obleser, Alfred Anwander, and Angela~D Friederici. 2012.
\newblock Linking ordering in broca's area to storage in left temporo-parietal
  regions: the case of sentence processing.
\newblock \emph{Neuroimage}, 62(3):1987--1998.

\bibitem[{Nivre(2004)}]{nivre2004incrementality}
Joakim Nivre. 2004.
\newblock Incrementality in deterministic dependency parsing.
\newblock In \emph{Proceedings of the workshop on incremental parsing: Bringing
  engineering and cognition together}, pages 50--57.

\bibitem[{Nivre(2008)}]{nivre2008algorithms}
Joakim Nivre. 2008.
\newblock Algorithms for deterministic incremental dependency parsing.
\newblock \emph{Computational Linguistics}, 34(4):513--553.

\bibitem[{Novick et~al.(2005)Novick, Trueswell, and
  Thompson-Schill}]{novick-cognitive-2005}
Jared~M. Novick, John~C. Trueswell, and Sharon~L. Thompson-Schill. 2005.
\newblock \href {https://doi.org/10.3758/CABN.5.3.263} {Cognitive control and
  parsing: {Reexamining} the role of {Broca}'s area in sentence comprehension}.
\newblock \emph{Cognitive, Affective \& Behavioral Neuroscience},
  5(3):263--281.

\bibitem[{Oh and Schuler(2023)}]{oh2023does}
Byung-Doh Oh and William Schuler. 2023.
\newblock Why does surprisal from larger transformer-based language models
  provide a poorer fit to human reading times?
\newblock \emph{Transactions of the Association for Computational Linguistics},
  11:336--350.

\bibitem[{Oota et~al.(2023)Oota, Marreddy, Gupta, and
  Bapi}]{oota-etal-2023-brain}
Subba~Reddy Oota, Mounika Marreddy, Manish Gupta, and Raju Bapi. 2023.
\newblock \href {https://doi.org/10.18653/v1/2023.findings-acl.415} {How does
  the brain process syntactic structure while listening?}
\newblock In \emph{Findings of the Association for Computational Linguistics:
  ACL 2023}, pages 6624--6647, Toronto, Canada. Association for Computational
  Linguistics.

\bibitem[{Pasquiou et~al.(2023)Pasquiou, Lakretz, Thirion, and
  Pallier}]{pasquiou2023information}
Alexandre Pasquiou, Yair Lakretz, Bertrand Thirion, and Christophe Pallier.
  2023.
\newblock Information-restricted neural language models reveal different brain
  regions' sensitivity to semantics, syntax and context.
\newblock \emph{arXiv preprint arXiv:2302.14389}.

\bibitem[{Pfeiffer et~al.(2020)Pfeiffer, R{\"u}ckl\'{e}, Poth, Kamath,
  Vuli\'{c}, Ruder, Cho, and Gurevych}]{pfeiffer2020AdapterHub}
Jonas Pfeiffer, Andreas R{\"u}ckl\'{e}, Clifton Poth, Aishwarya Kamath, Ivan
  Vuli\'{c}, Sebastian Ruder, Kyunghyun Cho, and Iryna Gurevych. 2020.
\newblock \href {https://www.aclweb.org/anthology/2020.emnlp-demos.7}
  {Adapterhub: A framework for adapting transformers}.
\newblock In \emph{Proceedings of the 2020 Conference on Empirical Methods in
  Natural Language Processing (EMNLP 2020): Systems Demonstrations}, pages
  46--54, Online. Association for Computational Linguistics.

\bibitem[{Prat and Just(2010)}]{prat10}
Chantel~S. Prat and Marcel~Adam Just. 2010.
\newblock \href {https://doi.org/10.1093/cercor/bhq241} {{Exploring the Neural
  Dynamics Underpinning Individual Differences in Sentence Comprehension}}.
\newblock \emph{Cerebral Cortex}, 21(8):1747--1760.

\bibitem[{Reddy and Wehbe(2021)}]{reddy21}
Aniketh~Janardhan Reddy and Leila Wehbe. 2021.
\newblock \href
  {https://proceedings.neurips.cc/paper_files/paper/2021/file/51a472c08e21aef54ed749806e3e6490-Paper.pdf}
  {Can fmri reveal the representation of syntactic structure in the brain?}
\newblock In \emph{Advances in Neural Information Processing Systems},
  volume~34, pages 9843--9856. Curran Associates, Inc.

\bibitem[{Roark(2001)}]{roark-2001-probabilistic}
Brian Roark. 2001.
\newblock \href {https://doi.org/10.1162/089120101750300526} {Probabilistic
  top-down parsing and language modeling}.
\newblock \emph{Computational Linguistics}, 27(2):249--276.

\bibitem[{Roark et~al.(2009)Roark, Bachrach, Cardenas, and
  Pallier}]{roark-etal-2009-deriving}
Brian Roark, Asaf Bachrach, Carlos Cardenas, and Christophe Pallier. 2009.
\newblock \href {https://aclanthology.org/D09-1034} {Deriving lexical and
  syntactic expectation-based measures for psycholinguistic modeling via
  incremental top-down parsing}.
\newblock In \emph{Proceedings of the 2009 Conference on Empirical Methods in
  Natural Language Processing}, pages 324--333, Singapore. Association for
  Computational Linguistics.

\bibitem[{Rodd et~al.(2010)Rodd, Longe, Randall, and
  Tyler}]{rodd2010functional}
Jennifer~M Rodd, Olivia~A Longe, Billi Randall, and Lorraine~K Tyler. 2010.
\newblock \href {https://doi.org/10.1016/j.neuropsychologia.2009.12.035} {The
  functional organisation of the fronto-temporal language system: evidence from
  syntactic and semantic ambiguity}.
\newblock \emph{Neuropsychologia}, 48(5):1324--1335.

\bibitem[{Schrimpf et~al.(2021)Schrimpf, Blank, Tuckute, Kauf, Hosseini,
  Kanwisher, Tenenbaum, and Fedorenko}]{schrimpf2020artificial}
Martin Schrimpf, Idan~Asher Blank, Greta Tuckute, Carina Kauf, Eghbal~A.
  Hosseini, Nancy Kanwisher, Joshua~B. Tenenbaum, and Evelina Fedorenko. 2021.
\newblock \href {https://doi.org/10/gncxrj} {The neural architecture of
  language: Integrative modeling converges on predictive processing}.
\newblock \emph{Proceedings of the National Academy of Sciences},
  118(45):e2105646118.

\bibitem[{Sedivy(2019)}]{sedivy19}
Julie Sedivy. 2019.
\newblock \emph{Language in Mind: An Introduction to Psycholinguistics}, second
  edition.
\newblock Oxford University Press.

\bibitem[{Shain et~al.(2020)Shain, Blank, van Schijndel, Schuler, and
  Fedorenko}]{shain2020fmri}
Cory Shain, Idan~Asher Blank, Marten van Schijndel, William Schuler, and
  Evelina Fedorenko. 2020.
\newblock {fMRI} reveals language-specific predictive coding during
  naturalistic sentence comprehension.
\newblock \emph{Neuropsychologia}, 138:107307.

\bibitem[{Sturt(1996)}]{sturt96}
Patrick Sturt. 1996.
\newblock \href {https://doi.org/10.1080/016909696387123} {Monotonic
  {S}yntactic {P}rocessing: {A} {C}ross-linguistic {S}tudy of {A}ttachment and
  {R}eanalysis}.
\newblock \emph{Language and Cognitive Processes}, 11(5):449--494.

\bibitem[{Vos et~al.(2001)Vos, Gunter, Schriefers, and Friederici}]{vos01}
Sandra~H. Vos, Thomas~C. Gunter, Herbert Schriefers, and Angela~D. Friederici.
  2001.
\newblock \href {https://doi.org/10.1080/01690960042000085} {Syntactic parsing
  and working memory: The effects of syntactic complexity, reading span, and
  concurrent load}.
\newblock \emph{Language and Cognitive Processes}, 16(1):65--103.

\end{thebibliography}
\bibliographystyle{acl_natbib}

\appendix

\clearpage

\onecolumn
\section{Appendix}
\label{sec:appendix}

\subsection{Generative Models}

As noted in Section \ref{sec:parser_training}, we use BLOOM's tokenizer to encode all tokens. Since some of these tokens are not found in the training data, and thus are unknown to the parser's next word prediction classifier, this prevents our parser from being strictly generative.

\begin{figure}[h]
    \centering
    \includeinkscape[width=0.5\columnwidth]{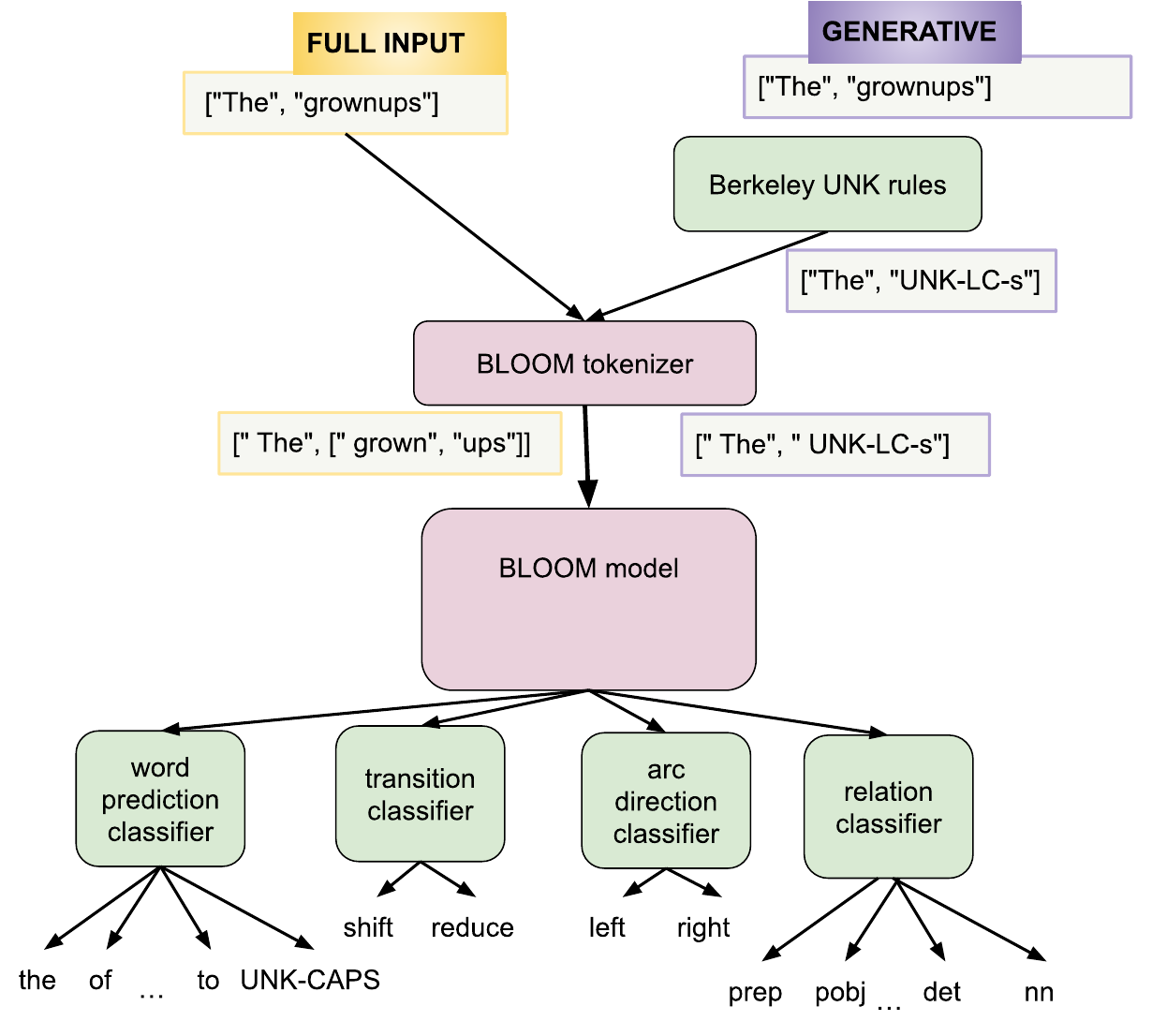}
    \caption{Parser input is pre-processed according to the given method: \textit{Full Input} indicates no pre-processing apart from adding a leading space to all pretokenized text from the corpus. \textit{Generative} indicates tokens are preprocessed by replacing tokens not seen more than once in the training data with unknown tokens according to the rules of the Berkeley Parser. The BLOOM model and the system of classifiers used to output probabilities is the same for both models. Note that the word prediction classifier predicts over the size of the training data vocabulary + Berkeley unknown tokens for both methods.}
    \label{fig:tokenization_flowchart_generative}
\end{figure}

As a comparison, we train additional models which are truly generative by replacing tokens not seen in the training data with unknown tokens in the encoder input, in the same way as the word prediction classes are defined. We report the accuracy for this \textit{generative} approach alongside the previously defined \mbox{\textit{full~input}} model in Table \ref{tab:parser_accuracy_generative}. This distinction is also explained in Figure~\ref{fig:tokenization_flowchart_generative} for an example prefix string. 

\begin{table}[h]
    \centering
    \scalebox{1}{
    \begin{tabular}{|c|c|c c|c c|}
        \hline
        \multirow{2}{*}{Corpus} & \multirow{2}{*}{Model} & \multicolumn{2}{c|}{Dev} & \multicolumn{2}{c|}{Test}  \\
        & & LAS & UAS & LAS & UAS  \\
        \hline
        PTB-3 & \citet{buys2018neural} & 88.66 & 91.19 & 88.54 & 91.01 \\
        PTB-3 & English BLOOM Generative & 89.27 & 92.00 & 90.53 & 92.68 \\ 
        PTB-3 & English BLOOM Full Input (brain analysis) & 90.26 & 92.71 &  90.32 & 92.62 \\
        \hline
        CTB-7 & Chinese BLOOM Generative & 73.12 & 80.60 & 74.47 & 81.60  \\
        CTB-7 & Chinese BLOOM Full Input (brain analysis) & 77.07 & 83.65 & 74.71 & 81.66 \\
        \hline
    \end{tabular}}
    \caption{Labeled attachment score (LAS) and unlabeled attachment score (UAS) for the English PTB corpus and Chinese CTB corpus}
    \label{tab:parser_accuracy_generative}
\end{table}

\subsection{Parser Training}

The output representation from the BLOOM model has a dropout of 0.5 applied, and is then fed into a single layer feedforward neural network. 

During training, gradient norms are clipped to 5.0, and the initial learning rate is 1.0, with a decay factor of 1.7 applied every epoch after the initial 6 epochs. English and Chinese models train for approximately 5.5 and 4.5 minutes, respectively, per epoch on an A100 GPU.

\subsection{Intermediate Results}

\begin{figure}[h]
    \centering
    \includegraphics[width=0.75\textwidth]{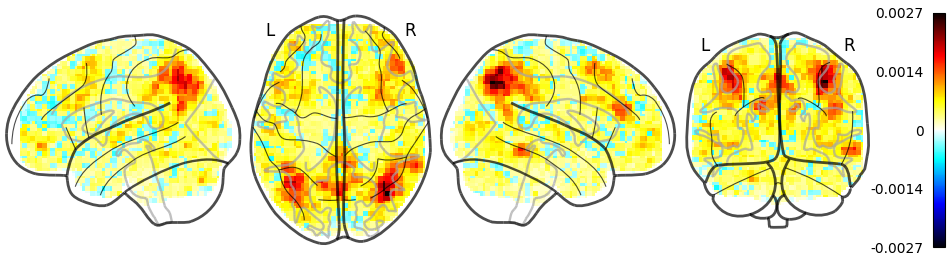}
    \caption{English Brain z-maps showing the $r^2$ increase for the model including syntactic~surprisal at $k=5$ compared to the model including only the following regressors: word rate, fundamental frequency (f0), word frequency, and root mean square intensity (RMS).}
    \label{fig:english_k5}    
\end{figure}
\begin{figure}[h]
    \centering
    \includegraphics[width=0.75\textwidth]{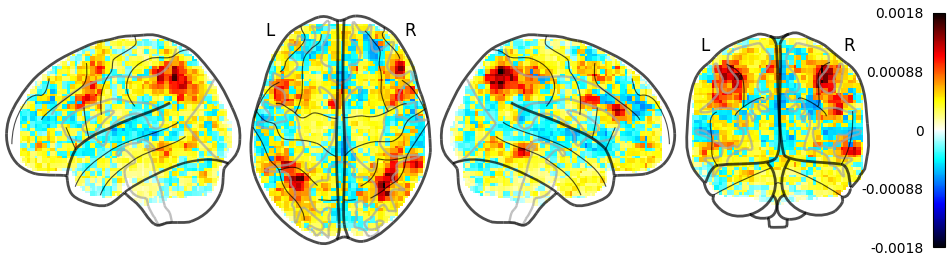}
    \caption{English Brain z-maps showing the $r^2$ increase for the model including syntactic~surprisal at $k=1$ compared to the model including only the following regressors: word rate, fundamental frequency (f0), word frequency, and root mean square intensity (RMS).}
    \label{fig:english_k1}    
\end{figure}

\begin{figure}[h]
    \centering
    \includegraphics[width=0.75\textwidth]{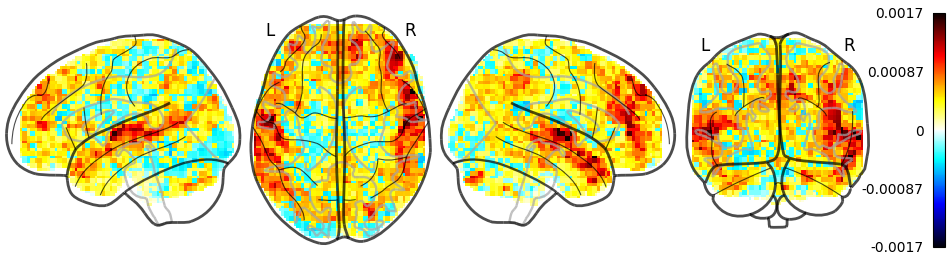}
    \caption{Chinese Brain z-maps showing the $r^2$ increase for the model including syntactic~surprisal at $k=5$ compared to the model including only the following regressors: word rate, fundamental frequency (f0), word frequency, and root mean square intensity (RMS).}
    \label{fig:chinese_k5}    
\end{figure}

\begin{figure}[h!]
    \centering
    \includegraphics[width=0.75\textwidth]{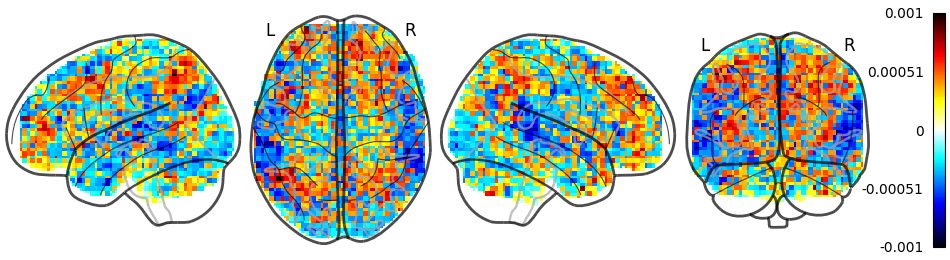}
    \caption{Chinese Brain z-maps showing the $r^2$ increase for the model including syntactic~surprisal at $k=1$ compared to the model including only the following regressors: word rate, fundamental frequency (f0), word frequency, and root mean square intensity (RMS).}
    \label{fig:chinese_k1}    
\end{figure}

\subsection{Results}

Tables \ref{tab:EN_results} and \ref{tab:CN_results} show the full list of significant clusters (p < .001 uncorrected; cluster threshold = 15 voxels) for the model comparison between syntactic surprisal at $k=1$ vs $k=5$.

\begin{table}[h]
    \centering
\begin{tabular}{|l|rrr|c|c|}
\hline
Region &     X &     Y &     Z &  Peak Stat & Cluster Size (mm3) \\
\hline
Left-SecVisual(18), Right/Left DorsalPCC(31) & -12.0 & -64.0 &  24.0 &       4.78 &                968 \\
Right-VisMotor(7) & 0.0 & -64.0 &  48.0 &       4.02 &                856 \\
Right-AngGyrus (39) & 38.0 & -54.0 &  38.0 &       4.42 &                800 \\
Left-PreMot + SuppMot(6) / Left-PrimAuditory (41) & -48.0 & -10.0 &   4.0 &       4.68 &               568 \\
Right-VentPostCing(23) & 14.0 & -54.0 &  20.0 &       4.57 &                488 \\
Right-AngGyrus (39) & 36.0 & -74.0 &  40.0 &       4.27 &                360 \\
Left-AngGyrus (39) & -44.0 & -68.0 &  36.0 &       3.77 &                352 \\
Right-DorsalPCC (31) & 14.0 & -62.0 &  34.0 &       4.03 &                216 \\
Right-DorsalPCC (31) & 4.0 & -50.0 &  44.0 &       3.65 &                192 \\
Right-PrimAuditory (41) & 48.0 & -12.0 &   2.0 &       3.69 &                160 \\
Right-PrimAuditory (41) & 38.0 & -26.0 &  12.0 &       3.39 &                152 \\
\hline
\end{tabular}    \caption{English significant clusters (p < .001 uncorrected; cluster threshold = 15 voxels) for the model comparison between syntactic surprisal at $k=1$ vs $k=5$}
    \label{tab:EN_results}
\end{table}

\begin{table}[h]
\begin{tabular}{|l|rrr|c|c|}
\hline
Region &     X &     Y &     Z &  Peak Stat & Cluster Size (mm3) \\
\hline
Right-MedTempGyrus (21) / Right-SupTempGyrus (22) &  56.0 & -36.0 &   4.0 &       4.68 &               1336 \\
Left-SupTempGyrus (22) & -64.0 & -26.0 &   6.0 &       4.32 &                392 \\
Right-PrimAuditory (41) & 52.0 & -14.0 &   6.0 &       4.59 &                384 \\
Left-PrimAuditory(41) & -56.0 &  -6.0 &   4.0 &       5.04 &                384 \\
Left-MedTempGyrus (21) & -48.0 & -42.0 &   4.0 &       4.25 &                320 \\
Left-SupTempGyrus (22) & -58.0 & -26.0 &   0.0 &       4.06 &                288 \\
\hline
\end{tabular}
\caption{Chinese significant clusters (p < .001 uncorrected; cluster threshold = 15 voxels) for the model comparison between syntactic surprisal at $k=1$ vs $k=5$}
\label{tab:CN_results}
\end{table}

\end{document}